\documentclass{article}
\usepackage{ijcai09}

\usepackage{times}

\usepackage[hyphens]{url}
\usepackage{hyperref}
\usepackage{breakurl}

\title{Semantic Robot Vision Challenge: Current State and Future Directions}
\author{Scott Helmer, David Meger, Pooja Viswanathan, Sancho McCann, \\ 
			{\bf Matthew Dockrey, Pooyan Fazli, Tristram Southey, Marius Muja,} \\
			{\bf Michael Joya, Jim Little, David Lowe, Alan Mackworth} \\
Department of Computer Science\\
University of British Columbia\\
shelmer@cs.ubc.ca}

\begin{document}
\sloppy

\maketitle

\begin{abstract}

The Semantic Robot Vision Competition provided an excellent opportunity for our research lab to integrate our many ideas under one umbrella, inspiring both collaboration and new research. The task, visual search for an unknown object, is relevant to both the vision and robotics communities. Moreover, since the interplay of robotics and vision is sometimes ignored, the competition provides a venue to integrate two communities. In this paper, we outline a number of modifications to the competition to both improve the state-of-the-art and increase participation.

\end{abstract}

\section{Introduction}

Current technology (robotic and otherwise) falls well short of a human's ability to perceive the world using vision. A nearly limitless range of applications would be facilitated by successful embodied object recognition (i.e., the ability of a mobile platform to perform human-like visual scene and object understanding). We believe that with several key advances in the ability of a computer system to interpret visual imagery, namely robust object recognition of a large number of object classes and more capable scene understanding, future robot systems will substantially enhance the lives of their users. A robot introduced into a home environment will quickly be able to respond to commands such as "Robot, fetch my shoes!", assistive mobility devices will be able to determine whether a dangerous object is in the user's path, and navigation systems will aid travelers by identifying accidents and construction delays.

In order to accelerate the progress of state-of-the-art research, many fields in science and engineering have employed standardized benchmarks or data sets to evaluate similar techniques and provide a means for their comparison. However, these measures can be detrimental when they do not reflect the reality or complexity of the problem in question. If the benchmark represents a severe simplification of reality, its use for evaluation of techniques may lead to overconfidence in a system's accuracy and robustness. In addition, they may discourage research directions that are not aligned with success on such benchmarks.

Research competitions, while potentially possessing these same limitations, are more desirable than standard benchmarks on many points. First, like standard benchmarks they provide a context in which participants can evaluate their techniques under uniform conditions and make meaningful comparisons. Second, their periodic nature allows the competition to evolve with the state-of-the-art, and discourages techniques tailored to specifics of a particular benchmark. Finally, they provide an exciting venue that brings together a community for collaboration and synthesis. However, this assumes that those engaged in the state-of-the-art research participate actively in such competitions, otherwise the events become merely a venue for displaying known techniques.

Although there have been competitions focused upon a variety of robotic tasks, these have tended to minimize the contribution of vision. Conversely, in vision, particularly object recognition, the active acquisition of images for analysis is generally of secondary concern. Here, benchmark datasets and competitions are neither a representative sample of the real world, nor a sample of how a robot would see the world. By separating robotics and vision, the majority of cutting-edge object recognition research has focused only upon appearance-based approaches, ignoring scene cues that may prove beneficial for both accuracy and efficiency. We believe that in order to push state-of-the-art methods towards the challenging goals outlined earlier in this paper, a competition must bring these communities together by evaluating embodied object recognition systems in realistic environments, and thus reducing over-simplifications and erroneous research directions.

A recent competition featuring embodied object recognition is the Semantic Robot Vision Challenge (SRVC). The overall task in this contest is similar to a photo-scavenger hunt in an unknown indoor environment, with information on the objects typically acquired from the Internet. This setting brings together numerous sub-fields of AI, including vision, robotics, and natural language processing, along with Internet search technologies. Although this competition does involve embodied vision and can help stimulate robotics and vision research, it has yet to gain notoriety in the research community and significantly advance the state-of-the-art.

Drawing upon our experience as a competitor in the SRVC for the past two years, we have identified issues in both robotic competitions and embodied recognition.  We provide an outlook for the future of the SRVC that will allow it to increase its impact on the community. Our contribution in this respect is two-fold. Firstly, we discuss the value of the existing SRVC competition to research in embodied vision, and how it has pushed our own research in new directions. Secondly, we review possible modifications to improve the competition in terms of the research directions it encourages and the number of participants it attracts.

\section{Robotics and Computer Vision Competitions}

Competitions in robotics and computer vision that display state-of-the-art techniques are a relatively recent phenomena. This is partially due to the fact that historically, the state-of-the-art in either domains were not mature enough to handle compelling tasks. However, beginning with Robocup, competitions have become a somewhat regular feature at both academic conferences and independent venues. It is worth taking a moment to consider some of the more successful competitions and the features that have made them relevant and viable.

The premier example of a successful competition is Robocup \cite{Kitano97}, pioneered by Alan Mackworth \cite{Mack93}, where robots compete against each other in a soccer-like setting. With an over-arching goal of having robots compete against humans in the mid-21 century, Robocup has proved to be a valuable education tool and testbed for many ideas in AI. One of the key features in its early success was that it offered a variety of leagues for participation. Robot and simulation leagues were offered, providing a venue for state-of-the-art research in robot control as well as techniques in planning and multi-agent systems. As a result, the competition has attracted a large number of participants, raising the profile of attendant research and providing a valuable research experience.

More recently, RoboCup@Home is a new RoboCup league which aims to develop service and assistive robots used in real-world personal domestic applications. The intent of the league is to promote the development of robotic technologies that can assist humans in everyday life. The competition proposes a number of benchmark tasks in a home environment, where success is determined by the number of tasks which the entrant's robot completes. Among one of the benchmarks is a task to find a specified object in the environment. Although this contest does contain some aspects of embodied vision, it does not offer a sufficiently challenging task to attract vision researchers to a competition that is not held in conjunction with AI or vision conferences.  It does, however, offer opportunities for teams to attempt a wide variety of tasks, each requiring expertise in different areas of research.  For example, while one task might require speech synthesis and aesthetic presentation, another might evaluate teams on safe navigation, tracking and human recognition.  This setup provides teams the flexibility to attempt specific tasks that they have research expertise in and opt out of others.

Another wildly successful competition in AI was the DARPA Grand Challenge \cite{gc2006}, offering one million dollars to the first team which could autonomously complete a 240 kilometer on- and off-road course. For the first Grand Challenge in 2004 the best competitor traveled just 11 kilometers before flipping over and catching on fire. In the following year, there were five vehicles which successfully completed the entire course. In 2007, just three and a half years after the first competition, six teams finished the Urban Challenge, which mixed robotic and non-robotic vehicles together in an urban setting and enforced California traffic laws. The Grand Challenge is the perfect illustration of a competition which pushed the state-of-the-art, particularly in systems engineering. Prior to the competition it was widely believed that current technology was simply not up to the challenge of this difficult task. This success was due in part to the fact that from the start it was well funded, attracted top-notch research institutions, received wide media attention, and provided a compelling task.

Another successful competition has arisen in the computer vision community, the Pascal Visual Object Classes Challenge (VOC) (\url{http://pascallin.ecs.soton.ac.uk/challenges/VOC/}). This is a EU-funded competition which began in 2005 with the goals of providing a yearly competition for object class recognition and localization and a set of standards and tools for evaluating algorithm performance. In contrast to standard benchmark datasets, entrants are evaluated on a novel dataset every year, which prevents algorithms from being tailored specifically to a single data set. The key feature of this competition that led to its success was that it involved high profile researchers at the organizing level, was held in conjunction with major conferences, and was relatively inexpensive to participants.

\section{SRVC and Our Experience}

Although the previously mentioned competitions have been successful, they do not address many of the issues of embodied vision. The SRVC is an ideal competition to push the state-of-the-art in this field. This competition was held for the first time at the Association for the Advancement of Artificial Intelligence (AAAI) conference in 2007 in Vancouver, and again in conjunction with the IEEE Conference on Computer Vision and Pattern Recognition (CVPR) 2008 in Alaska. The competition is a visual search task in an unknown environment. The entrants are given a list of objects to find in the environment, with the environment containing only a subset of those listed objects along with additional distractor objects. Using this list, the robots autonomously acquire data about these objects from the Internet in a fixed amount of time. Once data collection and learning are complete the robot searches the unknown environment with the task of finding the objects using the data acquired from the Internet. At the end of the exploration phase the robot returns an image for each object type containing a single bounding box around the target. The scoring is based on the bounding box accuracy. In addition, there is a software league, where the entrants are not responsible for acquiring images of the environment. Instead they are given a set of images taken of the environment that include both the objects and other scene elements.

Our team entered two versions of our Curious George robot to the 2007 and 2008 SRVC. We gained a wealth of experience during our system development process and actual contest participation, which will be described in the following section.

\subsection{Internet Data Collection and Filtering}
 
In the SRVC, all of the training data is acquired from the Internet at contest time with no human intervention. For visual appearance models, this generally means collecting a dataset of images via an Internet image search engine like Google. Given the varied nature of Internet image search results, a system was needed to filter the output before training could be performed. We implemented two phases of training data filtering. The first phase removed all cartoons, illustrations, technical schematics, and other non-photographic images using a quality score developed in \cite{quality2006}. The second phase prioritized groups of images that displayed a high degree of similarity, since we determined empirically that these images were more likely to contain the target object.

It is interesting to note that approaches for image filtering and ranking such as \cite{fergus04} are generally evaluated on a dataset drawn from a similar distribution to the training data. This is not the case for the SRVC scenario since data collected by a robot is not likely to be from canonical viewpoints in uncluttered backgrounds -- two properties that are common in Internet images. As a result, we did not pre-filter results based on generic object categorization techniques.

In addition to acquiring training images, other relevant data may be found on the Internet, such as contextual clues from LabelMe \cite{RussellIJCV2008} and size priors from the WalMart catalog. These all represent important sources of information for recognition purposes, but this potential has been left unaddressed in the literature. Although we were not able to integrate this information into our system in time for the SRVC competitions, they have encouraged us to pursue this direction for collecting additional training data aside form the traditional image datasets.

\subsection{Robotics}

The nature of a robotics contest demands the construction of a physical system with numerous abilities ranging from basic navigation, to the construction of a distributed computation system, to performance on the task itself (object recognition in our case). While each of these individual tasks are easily achieved, their integration within a physical system presents a high level of complexity for system designers. For example, from a research perspective, robot navigation and mapping is largely a solved problem in indoor environments. However, it is a significant practical challenge to prepare a robot to navigate a previously unseen contest environment where it is required to visit potentially unsafe locations that provide good views of objects. Similarly, distributing a computational process across several networked processors is not a significant challenge in many situations, but when this system must be mounted on a mobile robot and thus subject to constraints on weight, power and size, many difficulties present themselves.

For the 2008 SRVC, we developed an active exploration and real-time vision system in order to announce objects that were discovered in real-time during the run. Our robot's architecture involved the low power, on-board PC mounted inside our Pioneer AT3 robot for low-level control, and four networked laptop systems responsible for: i) real-time processing of visual imagery, visual attention, robotic planning, gaze planning, and overall control; ii) specific object recognition; iii+iv) generic category recognition. The distributed nature of this architecture required captured imagery to be transfered between computers via network connection, and the associated software for sending and receiving components.

Our visual attention system processed imagery obtained from a stereo camera system in real-time in order to determine the locations of interesting objects and structures in the environment. This represented a significant new functionality when compared with our 2007 contest entry that required the robot to ``stop and shoot'' before performing visual attention. The real-time functionality prevented our robot from ``driving blindly'' and allowed it to continuously monitor the peripheral view until a sufficiently interesting location was seen, at which point foveal images could be collected. This behaviour allowed the robot to cover the environment rapidly while ignoring uninteresting regions and thus capturing images of a large number of candidate objects. It is unlikely that any of our team members would have developed such a behaviour had it not been for the SRVC, since stationary visual attention is equally easy to demonstrate in academic publication. However, now that such a system has been developed, our team has the ability to evaluate the behaviour of interactive, real-time visual attention on a mobile platform, and this continues to be an interesting research direction for our research group.

\subsection{Vision}

Images collected by a robot during the embodied object recognition scenario often capture objects from a non-standard viewpoint, scale, or orientation. In other cases, the images do not contain an object at all. In fact, during our SRVC experience, we found that images collected by the robot rarely contained any target object. As a result, we designed our classification system to have a low false positive rate. We employed a two-stage object detection approach. The first stage used a specific object recognition system based on matching SIFT features and geometric consistency that generally produced few false positives but provided low recall for generic object classes. The second stage employed a generic object classifier based on the spatial pyramid match kernel \cite{Lazebnik} to produce detections for those objects that were not captured by the previous approach.

We designed a peripheral-foveal vision system that attempts to improve the quality of robot-collected imagery by locating interesting regions of the environment and imaging these regions in high resolution. This design choice was inspired by the human visual system, which makes extensive use of peripheral-foveal vision. Our peripheral camera was a Point Grey Research Bumblebee stereo camera with a relatively wide field of view. Spectral saliency \cite{hou_saliency_2007} was fused with stereo depth information to locate regions of interest in peripheral images. The foveal camera was a Canon G7 point-and-shoot camera. We employed the G7's high zoom, combined with a pan-tilt unit to obtain tightly cropped, high resolution images of interesting objects identified in the peripheral view. We found that the image quality obtained by our foveal system significantly improved object recognition performance. This is likely due to the fact that Internet images are also often captured by high-quality digital cameras.
 
\subsection{Benefits to Our Research}

UBC's participation in the SRVC has lead us to develop Curious George, a powerful evaluation platform that enabled further development of embodied recognition algorithms. In terms of quantifiable research output, the platform developed directly for the SRVC contest has lead to a number of publications \cite{meger07,MegerRAS2008} and several higher-level algorithms have since been designed which leverage the platform \cite{forssen08,Viswanathan09}. Our resulting research directions can be summarized into three categories: the effect of viewpoint in object recognition, the use of existing online databases for semantic training information, and the use of additional cues available to an embodied platform during scene understanding.

Our study of viewpoint in object recognition has examined the implications of having only a single canonical viewpoint in the training image dataset (as is often the case with Internet images). We evaluated several recognition methods (namely feature matching with and without a geometric constraint) in terms of their ability to recognize objects from a range of viewpoints, and reported a range of success for this task.

We showed that annotated datasets such as the LabelMe \cite{RussellIJCV2008} database can provide semantic information for tasks other than simply object recognition. Object-place relations from LabelMe (e.g., fridges are likely to be found in the kitchen) were learned, and used this spatial-semantic model to perform place labeling in simulated environments. We also described the use of this model to inform object search \cite{Viswanathan09}.  Our future plans are to combine this technology with object recognition, demonstrated in the SRVC, to construct a successful integrated scene understanding system.

Finally, we have employed structure from stereo to register object locations and construct a 3D object map and demonstrated how this object map allows a robot to collect multiple viewpoints of target objects to improve classification accuracy.  One of our team members is currently employing the raw structure information to utilize scale priors for object recognition. Overall, the SRVC has stimulated a wide variety of excellent research in our group by forcing us to examine object recognition in a realistic setting.

\section{Improving Research Outcomes}

Research competitions should advance the state-of-the-art by providing additional training data and context information.  In addition, realistic environments would allow the use of more advanced learning methods. This section provides potential modifications to the SRVC contest that we believe will encourage these directions.

\subsection{Training}
 
Embodied object recognition systems require a source of training data from which to learn the appearance and properties of target objects. In the past, for the SRVC, this data has been obtained entirely using the Internet at the time of competition. However, the vast majority of images from the Internet are from a single canonical viewpoint, which implies that the resulting classifier will only be successful on that viewpoint. Given the paucity of the data, this setting does not encourage 3D recognition, which may be required for successful embodied recognition. One modification would be to allow competitors to know a superset of the classes beforehand, enabling the use of manually labeled training data.  This is not an unrealistic scenario since most robots will likely be deployed in known environments where the set of objects can be carefully catalogued. Also, it still presents a significant challenge, as demonstrated by the VOC competition where recognition is still very poor. Alternatively, the types of environments (e.g., office, kitchen, bedroom, etc) could be provided.  For example, knowing that the scene was a kitchen would allow researchers to construct priors on appearance, 3d shape and scale for all objects that are likely to occur in a kitchen.  In either case, Internet data acquisition would still be allowed at competition time to augment data provided by system designers. This would allow for research into the interplay between scene information like surface orientations and real-world scale and appearance, similar to works such as \cite{hoeim06,gould08}.

\subsection{Environment and Context}

The SRVC contest environment has, so far, required the robot to navigate in an area that is quite small and to locate objects that were placed on tables covered with white table cloths. This scenario presents a much simpler segmentation problem when compared with a realistic home environment, and does not allow for evaluation of system performance over long distances and operating durations.

Thus, while object recognition methods that rely on good segmentation results might succeed in the contest, they are likely to fail in more realistic environments. This outcome is misaligned with the objective of pushing research in the direction of improving real-world performance, and we believe that future SRVC contests should include increasingly realistic environment designs.

To a na\"{\i}ve audience, embedding the competition in a realistic environment might seem likely to increase difficulty, however, it can actually lead to better performance if the additional information available about context is leveraged by the competition systems.  Respecting relationships of co-occurance and co-location of natural environments when placing objects would allow one to exploit these relationships for object recognition.  It would also help eliminate false matches by recognizing that an object does not belong in a particular location.

In addition, it would be interesting to partition the environment into places that appear in real environments (e.g., kitchen, bedroom, etc.) and having query objects in the locations that they are normally found.  This would allow competitors to exploit object-place relations to identify potential object locations, thus facilitating efficient coverage of the environment.  There are obviously logistical problems in having a multi-room environment and allowing for an audience.  However, using dividers, it is possible to create room-like subdivisions without the need for entirely separate rooms.  This would create an environment similar to many ``open concept'' homes.  It is also possible to create recognizable, logical locations in a single room by separating these locations with empty space, however this should be specified to competitors.

\section{Improving Participation}

The purpose of a competition in research is to provide both an opportunity to exchange ideas as well as a venue to evaluate and encourage state-of-the-art research. A particular challenge in an embodied recognition competition is to encourage participation of \emph{both} robotics and vision researchers.  In this section, we discuss practical suggestions to increase researcher participation.

\subsection{Changes in the Setting and Rules}

Various methods are currently used in object class recognition research such as colour, contours, texture, etc. In addition, there is an active research community \cite{vogel07} that seeks to utilize scene context for recognition. We propose varying the difficulty and scoring of the competition in a way that rewards the successes of specific methods on certain object types that might be challenging to recognize using simple object recognition techniques.

Another interesting modification might be to provide different levels of information before and during the competition. For example, the object type ``bottle'' could be provided beforehand, and the robot might be required to recognize a specific object (e.g. coke bottle, milk bottle, etc) during the competition. To make the problem more challenging, the contest could allot points for identifying unknown objects (i.e. those that do not appear on the list). In addition, including relative location information for some objects (e.g. the book is beside the TV) can provide context information useful for recognition of objects that are particularly challenging given the state-of-the-art.

Additionally, the contest could allow two teams to compete simultaneously. The team which finds the objects in the environment faster would receive a higher score. Another interesting case that can push forward the robotics aspect of the contest would be to allow multiple robots per team to explore the environment. The robots can cooperatively capture the images from different viewpoints and share the information to recognize the objects more precisely. However, this change is most likely infeasible to implement in the near future due to the complexity and cost of robots currently being used in SRVC.

\subsection{Software League}
\label{SoftwareLeague}
Although a competition which requires the integration of various research areas is desirable, such a competition discourages participation from smaller research groups that may not have the expertise to implement every aspect required for success. The software league is an example of separating the recognition task from the robotics challenges of active vision and navigation, however some modifications are needed in order to improve participation in this league.

The first thing to note is that the impact of this competition is dependent on the significance of the results in the competition. In the object recognition community, techniques are evaluated on a large number of images, thus ensuring that improvements over previous techniques are statistically significant. This is the case even in a competition environment like VOC. Results from the SRVC competition, however, carry little statistical significance due to the small sample size (e.g. one mug in the environment). One possibility to address this is in the software league. Here, image data can be acquired from real environments instead of the contest setting. This provides an opportunity to include much more realistic context. Images of the same target objects distributed in a natural environment, such as a kitchen or office, can be taken ahead of time. In order to incorporate the ``embodied vision'' aspect of the contest, additional information such as a map of the environment, the location and orientation of the camera for each image, and stereo image pairs can be provided with little extra effort. This would make the software league a more interesting research problem and help distinguish it from other object recognition competitions, thus attracting more participants. In addition, removing the limitations of data collection by a robot also allows for the creation of data sets composed of a larger number of objects and environments, thus increasing the statistical significance of the results.

\subsection{Robot League}

As already mentioned, it has been a significant challenge for teams in previous years of the SRVC contest to achieve reliable navigation within the contest environment. Since the primary research problems posed by the SRVC are not intended to focus on low-level robot navigation, it may be useful to consider relieving teams of the navigation burden in the future.

First, the contest organizers could provide entrants with a standardized robot platform that has basic navigation abilities. In this case, teams would only be responsible for higher level task planning and processing of the visual imagery obtained by the robot. While this solution solves many of the problems posed by navigation, it also unfortunately introduces several complications. Primarily, each team depends on a slightly different set of sensing modalities. During the 2007 and 2008 SRVC contests, we have seen: monocular video cameras, monocular still cameras, laser rangefinders, sonar range sensors, binocular stereo cameras, and multi-camera stereo systems. Any standardized test platform would be required to provide teams with some subset of these sensors, and this set would ideally be sufficiently large so that it does not discourage any teams from competing. Another significant challenge is the ability for each team to practice and develop on the standard platform. Either numerous platforms would need to be distributed, or teams would require periodic access to a single platform. Both of these options entail significant cost that would need to be minimized. This could likely be accomplished by employing a standard robot architecture such as ROS (\url{http://pr.willowgarage.com/wiki/ROS}) that would allow much of the development to occur in simulation and with surrogate robots for hardware testing.

A simpler method for reducing navigation challenges is to provide teams with a more detailed specification of the contest environment geometry. For example, knowing the exact size and shape of furniture allows for proper mounting of sensors and tuning of sensor models in mapping algorithms. In this case, it might be possible for each team to still employ their own robot.

\subsection{Facilitating Code Re-use}
Re-usable code is an important output of a successful competition, as it is in any collaborative effort. Since the goal of competitions is to move the state-of-the art in a desired direction, successive solutions to the competition's problem benefit from having previous work available as a starting point. Re-usable code also lowers the barrier to entry for teams new to the competition, enhancing accessibility of the competition, and in turn visibility.

Although code \emph{sharing} is encouraged/required by SRVC, subsequent re-use of this code appears to be non-existent. This is a result of the differences in platforms and approaches used by each of the participants. For example, our robot base, sensor package, peripheral-foveal vision system, and multi-processor distributed recognition system was a very specific point in the solution space. This entire setup would likely have to be replicated in order to re-use our code. However, there are elements of a code base that would be generally usable (training set construction and feature extraction, for example).

One possible solution would be to require the use of an open source robotics package with a distributed architecture such as ROS, which allows different components to be easily chained together. In such a system, the different components are unaware of each other, so they can be mixed-and-matched at will. Standardization on a single robotics platform would be an optimal solution if the funding were available.

Without the logistical problems inherent in hardware, the software league offers much greater potential for code sharing. One possibility to help encourage this would be to design the software challenges to be explicitly modular in nature. Instead of a single software league challenge, it could be broken into steps such as download and filtering, classification, and localization. 
 
\subsection{Funding and Visibility}

A significant constraint for both organization and participation is funding. It requires a large amount of exhibition space in which to set up the environment. Obstacles and target objects for the environment must be purchased. Support must be offered to teams to subsidize the cost of shipping their robots to the competition in order to encourage participation. In addition, travel costs for the robot teams can be high since a large number of team members may be needed to run and maintain the robotic hardware. Clearly, secure funding through public and private sponsorship will attract participation. 

Aside from attracting more participation from research groups, improving the visibility of the contest can attract sponsorship. One simple change to the 2008 competition that was surprisingly compelling was the addition of bonus points for teams who made a realtime status display showing matches as they were made. This made the contest more interesting for the audience to watch by providing a sense of what the robots were doing even when they were not moving. The crowd was audibly excited when a new match was displayed, identifying with the robots and responding to the irregular reinforcement aspect of the display. As an element of visibility and outreach for a robotic competition, this is a very powerful lesson. In addition, this also pushed research towards techniques to provide real-time recognition. In future, explicitly encouraging competitors to provide real-time displays of what the robot has found or is trying to do will draw even more attention to the contest.

\section{Conclusions}
 
Properly designed contests significantly promote the development of the state-of-the-art. They can comprise realistic and complex settings not seen in standard benchmark datasets, providing both a strong test for current solutions and rich context that can be leveraged to advance research. The Semantic Robot Vision Challenge represents one such competition which provides a venue for embodied recognition. It has provided a valuable impetus to our own research, providing insights and elucidating new directions that need more research. However, to be successful in the future, this contest needs numerous modifications in order to have significant impact.


\bibliographystyle{named}
\bibliography{ijcai09}

\begin{thebibliography}{}

\bibitem[\protect\citeauthoryear{Fergus \bgroup \em et al.\egroup
  }{2004}]{fergus04}
R.~Fergus, P.~Perona, and A.~Zisserman.
\newblock A visual category filter for google images.
\newblock In {\em Proceedings of the 8th European Conference on Computer
  Vision, Prague, Czech Republic}, pages 242--256, May 2004.

\bibitem[\protect\citeauthoryear{Forssen \bgroup \em et al.\egroup
  }{2008}]{forssen08}
P.~E. Forssen, D.~Meger, K.~Lai, S.~Helmer, J.~J. Little, and D.~G. Lowe.
\newblock Informed visual search: Combining attention and object recognition.
\newblock In {\em Proceedings of ICRA}, May 2008.

\bibitem[\protect\citeauthoryear{Gould \bgroup \em et al.\egroup
  }{2008}]{gould08}
Stephen Gould, Paul Baumstarck, Morgan Quigley, , Andrew~Y. Ng, and Daphne
  Koller.
\newblock Integrating visual and range data for robotic object detection.
\newblock In {\em ECCV Workshop on Multi-camera and Multi-modal Sensor Fusion
  Algorithms and Applications (M2SFA2)}, 2008.

\bibitem[\protect\citeauthoryear{Hoiem \bgroup \em et al.\egroup
  }{2006}]{hoeim06}
D.~Hoiem, A.A. Efros, and M.~Heber.
\newblock Putting objects in perspective.
\newblock In {\em CVPR}, 2006.

\bibitem[\protect\citeauthoryear{Hou and Zhang}{2007}]{hou_saliency_2007}
Xiaodi Hou and Liqing Zhang.
\newblock Saliency detection: A spectral residual approach.
\newblock {\em Computer Vision and Pattern Recognition, IEEE Computer Society
  Conference on}, 0:1--8, 2007.

\bibitem[\protect\citeauthoryear{Ke \bgroup \em et al.\egroup
  }{2006}]{quality2006}
Yan Ke, Xiaoou Tang, and Feng Jing.
\newblock The design of high-level features for photo quality assessment.
\newblock In {\em CVPR '06: Proceedings of the 2006 IEEE Computer Society
  Conference on Computer Vision and Pattern Recognition}, pages 419--426,
  Washington, DC, USA, 2006. IEEE Computer Society.

\bibitem[\protect\citeauthoryear{Kitano \bgroup \em et al.\egroup
  }{1997}]{Kitano97}
Hiroaki Kitano, Minoru Asada, Yasuo Kuniyoshi, Itsuki Noda, and Eiichi Osawa.
\newblock Robo{C}up: {T}he robot world cup initiative.
\newblock In W.~Lewis Johnson, editor, {\em Proceedings of the First
  International Conference on Autonomous Agents}, New York, 1997. ACM Press.

\bibitem[\protect\citeauthoryear{Lazebnik \bgroup \em et al.\egroup
  }{2006}]{Lazebnik}
S.~Lazebnik, Cordelia Schmid, and Jean Ponce.
\newblock Beyond bags of features: Spatial pyramid matching for recognizing
  natural scene categories.
\newblock In {\em CVPR}, 2006.

\bibitem[\protect\citeauthoryear{Mackworth}{1993}]{Mack93}
Alan~K. Mackworth.
\newblock On seeing robots.
\newblock In A.~Basu and X.~Li, editors, {\em Computer Vision: Systems, Theory
  and Applications}, pages 1--13. World Scientific Press, Singapore, 1993.
\newblock Reprinted in P. Thagard (ed.), Mind Readings, MIT Press, 1998.

\bibitem[\protect\citeauthoryear{Meger \bgroup \em et al.\egroup
  }{2007}]{meger07}
David Meger, Per-Erik Forss{\'e}n, Kevin Lai, Scott Helmer, Tristram~Southey
  Sancho~McCann, Matthew Baumann, James~J. Little, David~G. Lowe, and Bruce
  Dow.
\newblock Curious george: An attentive semantic robot.
\newblock In {\em IROS 2007 Workshop: From sensors to human spatial concepts},
  San Diego, CA, USA, November 2007. IEEE.

\bibitem[\protect\citeauthoryear{Meger \bgroup \em et al.\egroup
  }{2008}]{MegerRAS2008}
D.~Meger, P.-E. Forssén, K.~Lai, S.~Helmer, S.~McCann, T.~Southey, M.~Baumann,
  J.~J. Little, and D.~G. Lowe.
\newblock Curious george: An attentive semantic robot.
\newblock {\em Robotics and Autonomous Systems Journal, Special Issue From
  Sensors to Human Spatial Concepts}, June 2008.

\bibitem[\protect\citeauthoryear{Russell \bgroup \em et al.\egroup
  }{2008}]{RussellIJCV2008}
B.~Russell, A.~Torralba, K.~Murphy, and W.~Freeman.
\newblock Labelme: a database and web-based tool for image annotation.
\newblock {\em International Journal of Computer Vision (special issue on
  vision and learning)}, 77(1-3):157 -- 173, 2008.

\bibitem[\protect\citeauthoryear{Seetharaman \bgroup \em et al.\egroup
  }{2006}]{gc2006}
Guna Seetharaman, Arun Lakhotia, and Erik~Philip Blasch.
\newblock Unmanned vehicles come of age: The darpa grand challenge.
\newblock {\em Computer}, 39(12):26--29, 2006.

\bibitem[\protect\citeauthoryear{Viswanathan \bgroup \em et al.\egroup
  }{2009}]{Viswanathan09}
Pooja Viswanathan, David Meger, Tristram Southey, James~J. Little, and Alan
  Mackworth.
\newblock Automated spatial-semantic modeling with applications to place
  labeling and informed search.
\newblock In {\em Proceedings of Canadian Robot Vision}, Kelowna, Canada, 2009.

\bibitem[\protect\citeauthoryear{Vogel and Murphy}{2007}]{vogel07}
Julia Vogel and Kevin Murphy.
\newblock A non-myopic approach to visual search.
\newblock In {\em Proceedings of the Fourth Canadian Conference on Computer and
  Robot Vision CRV}, Montreal, Canada, May 2007.

\end{thebibliography}

\end{document}